\documentclass{sig-alternate}

\usepackage{graphicx}
\usepackage{subfigure}
\usepackage{multirow}
\usepackage{algorithm}
\usepackage{algorithmic}
\usepackage{times}
\usepackage{paralist}
\usepackage{array}
\usepackage{authblk}
\usepackage{epstopdf}
\usepackage{hyperref}

\begin{document}

\conferenceinfo{RecSys 2014, 2nd Workshop on Large Scale Recommender Systems}{}
\CopyrightYear{2014}
\crdata{0-89791-88-6/97/05}
\clubpenalty=10000
\widowpenalty=10000

\title{Large Scale Purchase Prediction with \\ Historical User Actions on B2C Online Retail Platform}

\author[ \; ,1]{Yuyu Zhang\thanks{These authors equally contributed to this work.}}
\author[,1]{Liang Pang$^*$}
\author[,2]{Lei Shi$^*$}
\author[ \; ,1]{Bin Wang\thanks{Team advisor}}
\affil[1]{Institute of Computing Technology, Chinese Academy of Sciences}
\affil[2]{Institute of Software, Chinese Academy of Sciences}
\affil[ ]{\{zhangyuyu2008, pl8787, shilei1025\}@gmail.com, wangbin@ict.ac.cn}

\maketitle

\begin{abstract}
    This paper describes the solution of \emph{Bazinga} Team for Tmall Recommendation Prize 2014. With real-world user action data provided by Tmall, one of the largest B2C online retail platforms in China, this competition requires to predict future user purchases on Tmall website. Predictions are judged on $F_1Score$, which considers both precision and recall for fair evaluation. The data set provided by Tmall contains more than half billion action records from over ten million distinct users. Such massive data volume poses a big challenge, and drives competitors to write every single program in MapReduce fashion and run it on distributed cluster. We model the purchase prediction problem as standard machine learning problem, and mainly employ regression and classification methods as single models. Individual models are then aggregated in a two-stage approach, using linear regression for blending, and finally a linear ensemble of blended models. The competition is approaching the end but still in running during writing this paper. In the end, our team achieves $F_1Score$ 0.0611 and ranks 7th (out of 7,276 teams in total).
\end{abstract}


\vspace{10pt}

\section{Introduction}\label{sec:intro}

Tmall is a Chinese-language website for business-to-consumer (B2C) online retail, spun off from Taobao and operated in China by Alibaba Group.
For B2C platform like Tmall, both sellers and buyers play the role of its customers, which should be well served at the same time.
In other words, B2C platform should help sellers (brands) to precisely target potential buyers, and meanwhile help buyers (users) to easily find what they like to buy.
In that sense, purchase prediction or brand recommendation is crucially important for B2C online retail platforms.
On popular B2C platforms, e.g. Amazon or Tmall, the amount of active brands and users is huge.
For these B2C platforms, it is challenging to build large scale recommender system.
So here comes Tmall Recommendation Prize 2014, a large scale competition (over 7,000 teams participated) for large scale recommender system.
Tmall provides real-world business data containing more than half billion action records from over ten million distinct users.
To our best knowledge, this data set is the biggest ever among all the public competitions on recommender system.
Such big data poses a big challenge.
Luckily, a shared powerful cluster and corresponding development environment is provided for free during the competition.
Competitors are required to implement their ideas, even the simplest ones, in MapReduce fashion and ensure programs can correctly run on distributed cluster.
Due to the length limitation, details of distributed implementation will be omitted in this paper.

The competition task is clear and simple: recommend a set of brands to a set of users.
The provided data set consists of a number of user action records, following the schema as (user ID, brand ID, action type, date).
All the IDs of users and brands are hash-mapped and encrypted.
There are altogether 4 different types of user action: ``click'', ``buy'', ``collect'', and ``cart''.
No explicit training and testing set provided.
The competition requires to predict what will those users buy in the next month.
To be more specific, it requires to predict a set of (user ID, brand ID) pairs which may have ``buy'' records (no matter how many times) during the next month.
Since there is no explicit testing set, predictions are judged on $F_1Score$ for fair evaluation, which is defined based on $Precision$ and $Recall$:

\begin{equation}
    Precision = \frac{\sum_{i}^{N}hitBrand_i}{\sum_{i}^{N}pBrand_i},
\end{equation}

\begin{equation}
    Recall = \frac{\sum_{i}^{M}hitBrand_i}{\sum_{i}^{M}bBrand_i},
\end{equation}

\begin{equation}
    F_1Score = \frac{2 * Precision * Recall}{Precision + Recall},
\end{equation}
where $N$ stands for the number of users in prediction,
$M$ stands for the actual number of buyers in answer,
$hitBrand_i$ is the number of correctly predicted brands for user $i$,
$pBrand_i$ is the number of predicted brands for user $i$,
and $bBrand_i$ is the actual number of brands purchased by user $i$ in answer.

To discourage overfitting, a whole new data set (drawn from the same distribution) is scheduled to be released at the last week of competition.
The original data set will be replaced at that time.
Since the paper deadline is earlier than the end of this competition, this solution paper is all based on the original data set.

The remainder parts of this paper are organized as follows.
Section~\ref{sec:data} presents our basic data analysis and preprocessing approach.
In Section~\ref{sec:modeling}, we introduce our modeling on the purchase prediction task.
Section~\ref{sec:feature} describes our feature design.
Blending and ensemble algorithms are presented in Section~\ref{sec:blending}.
At last, we conclude the paper in Section~\ref{sec:conclusion}.

\section{Data Preprocessing}\label{sec:data}

In this section, we firstly conduct basic data analysis, and accordingly introduce data splitting for offline validation, as well as data cleansing.

    \subsection{Data Analysis}\label{sec:data_analysis}

    In the data set, record date ranges from April to July (from 04-15 to 08-15, year unknown).
    For convenience, we have a month name convention for the 4-month long data, as shown in Table~\ref{table:date}.

    \begin{table}
    \centering
    \caption{Month name convention.}\label{table:date}
    \vspace{10pt}
    \begin{tabular}{ccc}
      \hline
      Month & From Date & To Date \\
      \hline \hline
      April & 04-15 & 05-16 \\
      May & 05-17 & 06-20 \\
      June & 06-21 & 07-18 \\
      July & 07-19 & 08-15 \\
      \hline
    \end{tabular}
    \end{table}

    Table~\ref{table:stats} shows the basic statistics of the data set.
    From this table, we see that the data is almost uniformly distributed in those 4 months.
    On the average, each user has about 50 action records, while each brand has about 20,000 action records.
    For the brand level aggregation, the comparatively abundant information can be very helpful.
    As for the four action types, it is obvious that click is the ``cheapest'' one.
    The overall ratio between click and buy is about 39:1.

    \begin{table*}
    \centering
    \caption{Basic statistics of the data set.}\label{table:stats}
    \vspace{10pt}
    \begin{tabular}{lllll|l}
      \hline
       & April & May & June & July & \textbf{Date Total} \\
      \hline \hline
      (User ID, Brand ID) & 52,909,766 & 59,299,203 & 49,776,687 & 50,188,852 & 195,303,359 \\
      User ID & 7,513,602 & 8,024,680 & 7,415,736 & 7,497,824 & 12,500,984 \\
      Brand ID & 21,305 & 23,779 & 25,742 & 28,036 & 29,706 \\
      \hline
      Action Click & 133,211,931 & 153,814,233 & 134,127,435 & 128,902,358 & 550,055,957 \\
      Action Buy & 3,441,803 & 3,823,481 & 3,483,978 & 3,345,793 & 14,095,055 \\
      Action Collect & 1,727,565 & 1,740,057 & 1,452,666 & 1,576,096 & 6,496,384 \\
      Action Cart & 255,432 & 312,437 & 313,465 & 377,750 & 1,259,084 \\
      \hline
      \textbf{Action Total} & 138,636,731 & 159,690,208 & 139,377,544 & 134,201,997 & 571,906,480 \\
      \hline
    \end{tabular}
    \end{table*}

    \subsection{Data Splitting}\label{sec:splitting}

\begin{figure}
    \centering
    \includegraphics[width=.47\textwidth]{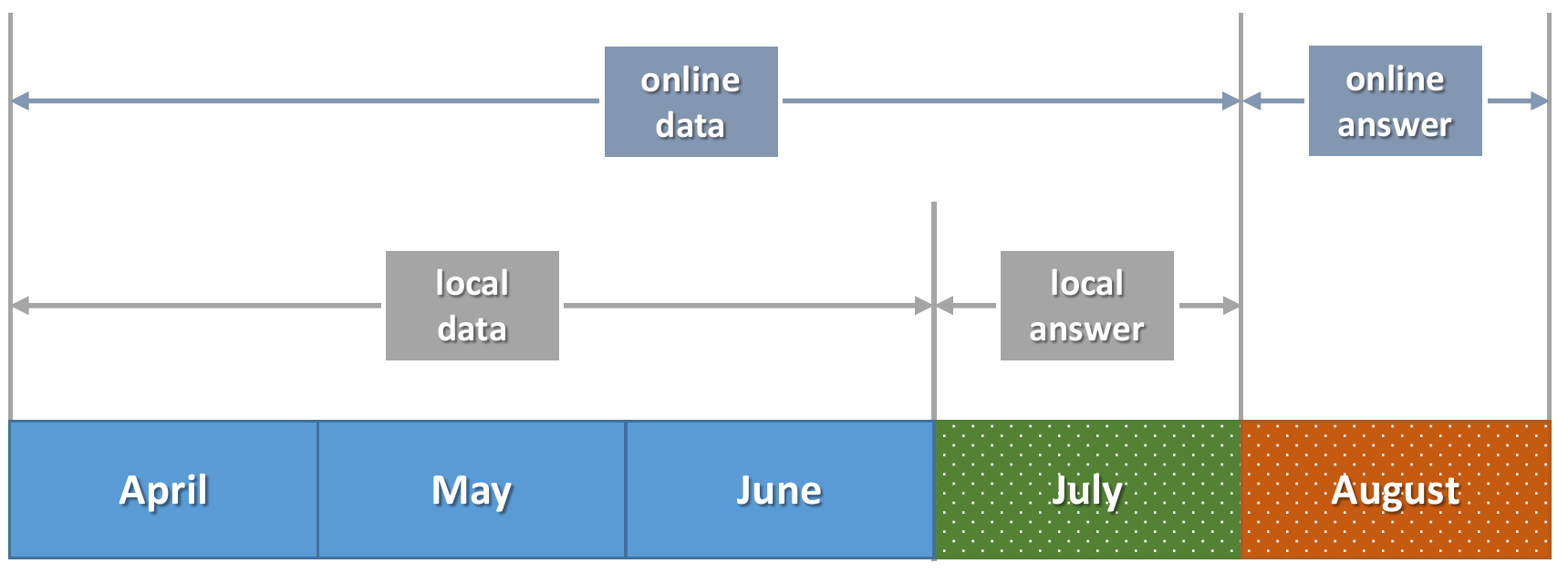}
    \caption{Data splitting for local validation.} \label{fig:data_split}
    \end{figure}

    Based on previous data analysis, we split the data set for local validation.
    Figure~\ref{fig:data_split} illustrates the details.
    For online setting, provided data set locates in the first 4 months, and the August data is an invisible answer set used for evaluating predictions.
    As for local data splitting, we use records in the first three months (April, May, and June) as the local data, and use records in July as the local answer set, which is invisible and used for evaluating local predictions.

    \subsection{Data Cleansing}\label{sec:clean}

    Data cleansing is necessary since we have observed some users have millions of clicks but no purchases.
    A possible explanation is that these users may be crawlers, which automatically access to target website many times every day.
    Obviously, such ``user'' should not be considered in model.
    We use a simple filtration rule to clean such noise in data: if a user have more than 500 clicks but 0 buys, then all the action records of this user are removed.
    According to our statistics, about 22 million action records are filtered out, which account for 4\% of the data set.

\section{Purchase Prediction Modeling}\label{sec:modeling}

In this section, we introduce the modeling of purchase prediction problem, including how to build instance and how to design time spans of feature and target.

    \subsection{Instance Building}\label{subsec:instance}

    We model the purchase prediction task as standard machine learning problem.
    We take (user ID, brand ID) pair as the instance.
    User actions are obviously non-i.i.d. and highly dependent on historical actions.
    For example, user is much more likely to purchase a certain brand if she has clicked and added it to shopping cart before.
    Therefore, we choose to build instance in a time-dependent fashion: the target of instance is decided by future actions.
    Here target stands for the ground truth of instance.
    If the task is modeled as a classification problem, the target is a binary label of purchase or not in the future.
    While if modeled as regression problem, the target can be either a binary label or the purchase times in the future.

\begin{figure}
    \centering
    \includegraphics[width=.47\textwidth]{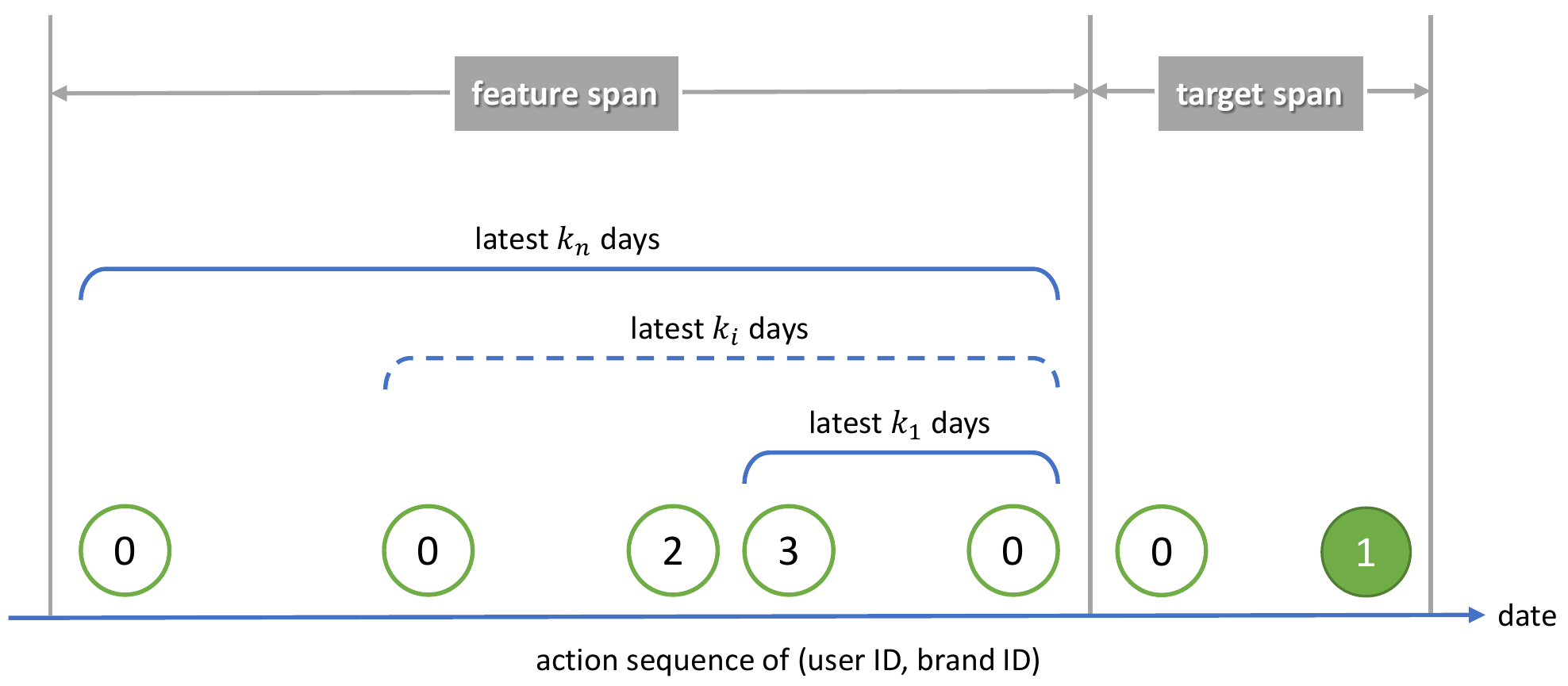}
    \caption{Illustration of instance building.} \label{fig:instance}
    \end{figure}

    Take classification modeling as an example, for each (user ID, brand ID) pair, what we need to predict is a conditional probability:
    \begin{equation}
        P(purchase\ in\ next\ month\ |\ historical\ actions),
    \end{equation}
    where $histrocial\ actions$ are the action records in previous months, including the actions of other users and brands.
    We use the given historical actions to build features of each instance, and set the corresponding target by checking whether or not the instance has any purchase action in next month.
    So that for each instance (user ID, brand ID), the records in action sequence are then divided into two parts by date: feature span and target span, as illustrated in Figure~\ref{fig:instance}.
    In this figure, digit 0-3 stands for click, buy, collect and cart action respectively.
    We build features within a set of date buckets, e.g. \{latest $k_1$ days, latest $k_2$ days, ..., latest $k_n$ days\}.
    Details of feature design will be discussed in Section~\ref{sec:feature}.

    \subsection{Time Span}\label{subsec:time_span}

\begin{figure}
    \centering
    \includegraphics[width=.47\textwidth]{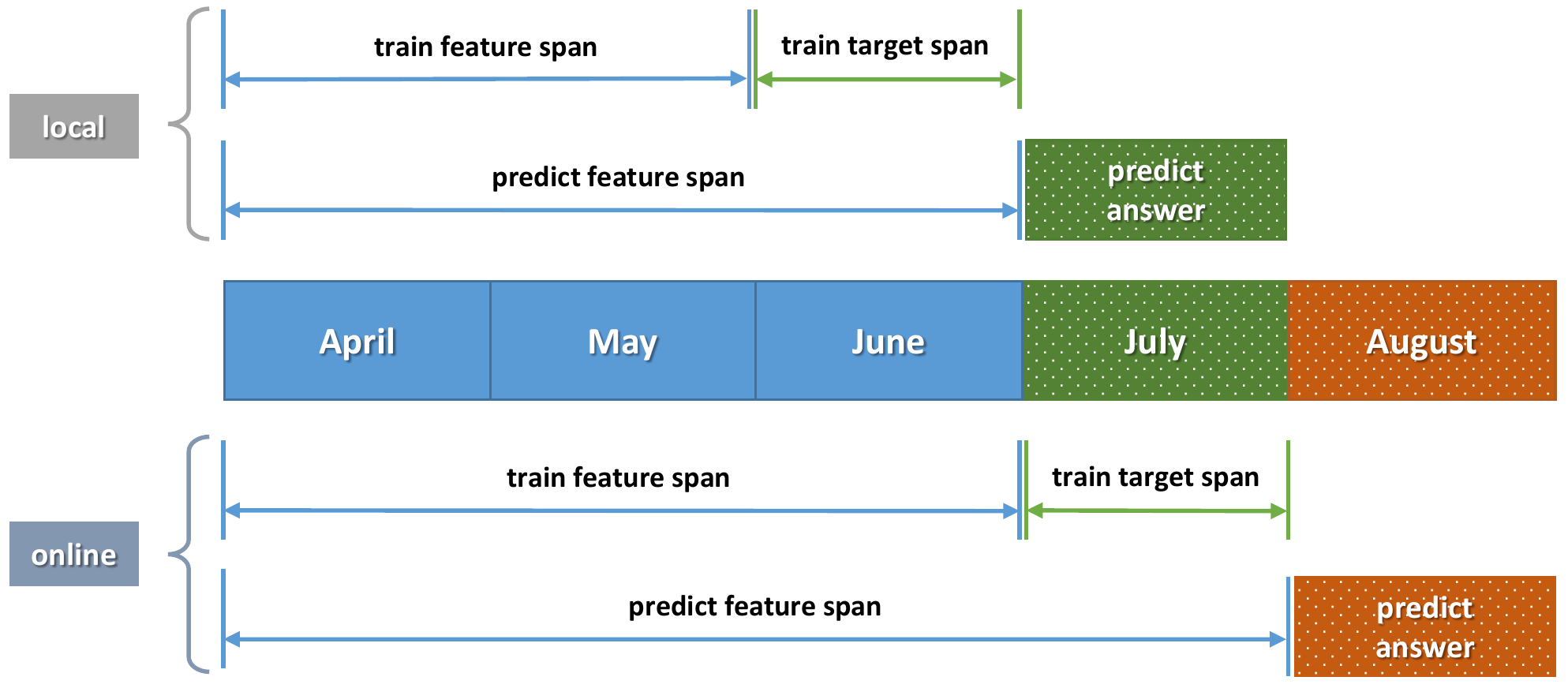}
    \caption{Fixed time spans on local and online data.} \label{fig:fixed_span}
    \end{figure}

\begin{figure}
    \centering
    \includegraphics[width=.47\textwidth]{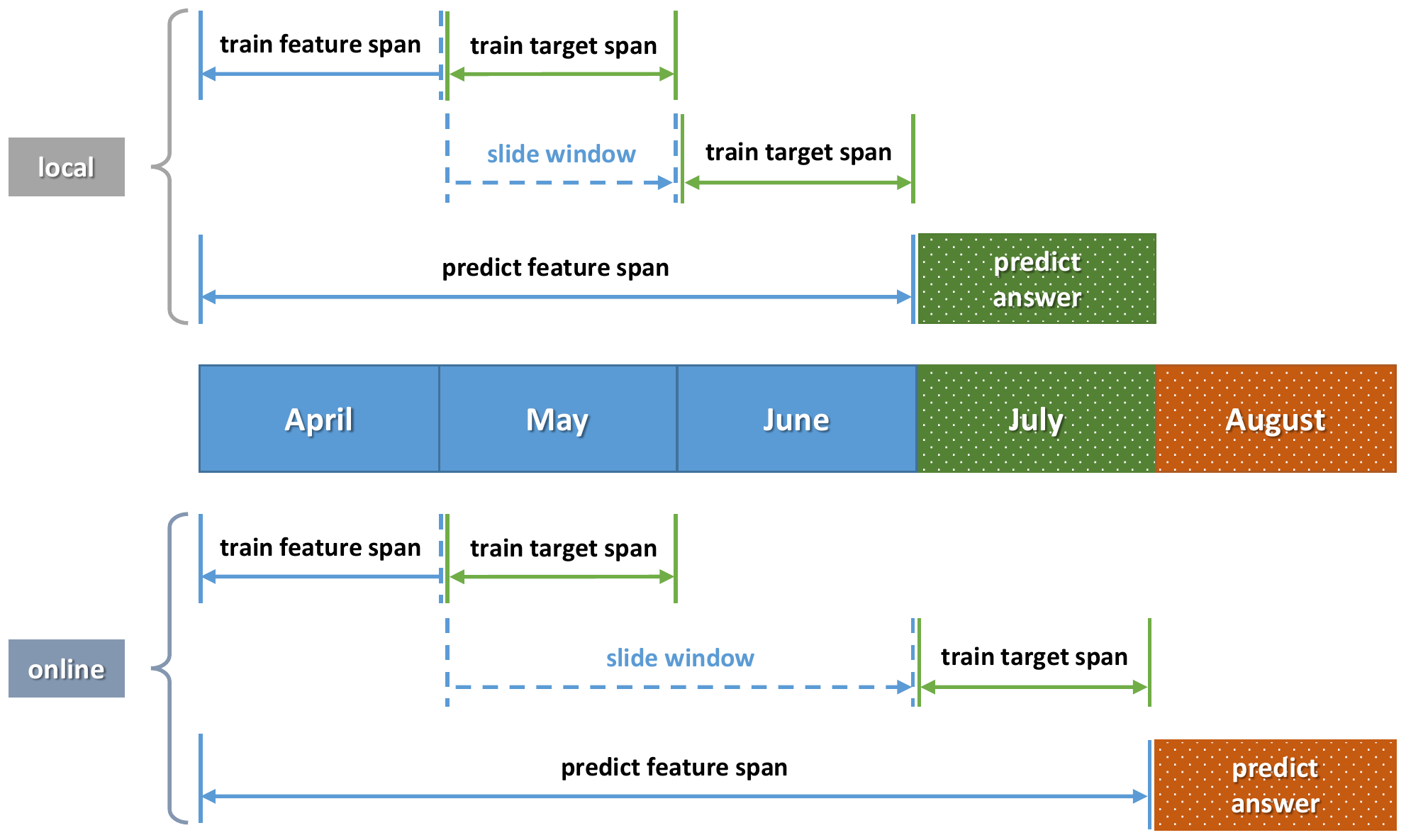}
    \caption{Sliding time spans on local and online data.} \label{fig:sliding_span}
    \end{figure}

    As discussed above, the time spans of feature and target are separated.
    We have two different ways to construct time spans:
    \begin{itemize}
      \item Fixed time spans. In training, we use the last month as the target span and previous months as the feature span. In predicting, we use all available months as the feature span. Figure~\ref{fig:fixed_span} illustrates the details of fixed time spans in both local and online settings.
      \item Sliding time spans. In training, the feature span becomes extendable, and the target span follows the sliding end of the feature span. The length of target span is still fixed as one month. In predicting, it keeps the same as fixed time spans. Figure~\ref{fig:sliding_span} illustrates the details of sliding time spans in both local and online settings.
    \end{itemize}

\section{Feature Design}\label{sec:feature}

In this section, we introduce our design of features.
Basically, for an instance (user ID, brand ID) pair, we build its features in three parts: pair features, user features, and brand features.
For each part, we design 5 groups of features: counting features, ratio features, flag features, and global features.
We will list each feature which parts it is applied to in detailed feature list as follows.

    \subsection{Counting Features}\label{subsec:counting_features}

    We use several types of counting features as the basic statistics on various granularities, which are listed below.

    \begin{itemize}
      \item Action count, including the count of each action type and the sum of all action types. This counting feature is applied to pair, user and brand.
      \item Action day count, including the day count of each action type. This is an aggregation of the action count on the day level. This counting feature is applied to pair, user and brand.
      \item Valid click count, which only counts the number of clicks later than the last purchase. This counting feature is only applied to pair.
      \item Distinct action user or brand count, which counts the number of distinct users or brands in each action type. This counting feature is applied to brand and user.
      \item First-time action user or brand count, which counts the number of initial action of user or brand. This counting feature is applied to brand and user.
    \end{itemize}

    \subsection{Ratio Features}\label{subsec:ratio_features}

    Ratio features are generated by division between counting features.
    We mainly build two types of ratio features.

    \begin{itemize}
      \item Conversion ratio, which divides purchased related count (buy count, buy day count) by the total action count. Conversion ratio reflects the likelihood of purchase after click. This ratio feature is applied to pair, user and brand.
      \item Counting ratio, which divides click or buy count by another type of click or buy count. Counting ratio is between counting features of the same action type. For example, click count / click day, buy count / distinct buy user, etc. This ratio feature is applied to user and brand.
      \item Cross ratio, which is built between pair and user counting features. For example, user click count / pair click count. Cross ratio shows the user's preference on specific brand in pair.
    \end{itemize}

    \subsection{Flag Features}\label{subsec:flag_features}

    Rules might be straightforward but very useful as additional predictions, which can be merged into model predictions.
    For example, a simple rule can be just predicting pairs with more than 10 clicks.
    However, our offline experiments show that it doesn't work if directly add rule predictions on model output.
    Therefore, we transfer effective rules to binary flags and elegantly incorporate flag features in learning model, which works better than directly adding rule predictions.
    Our flag features include:
    \begin{itemize}
      \item Whether or not this pair or user or brand has a certain type of action before. This flag feature is applied to pair, user and brand.
      \item Whether or not this pair or user has consecutive purchases in adjacent two months. This flag feature is applied to pair, user and brand.
    \end{itemize}

    \subsection{Global Features}\label{subsec:global_features}

    Counting, ratio, and flag features are all built based on date buckets, i.e. each feature has a copy in each date bucket and the feature value is determined by the range of date bucket.
    We also build global features, which are not dependent on date buckets and computed on the whole feature span.
    Global features have more information than date bucket dependent features. They are more robust and can describe instance more precisely.
    We mainly use the following global feature:
    \begin{itemize}
      \item First and last active day, which is the distance between the first or last active day and the end of the feature span. This global feature is applied to pair and user.
      \item Last purchase day, which is the distance between the last purchase day and the end of the feature span. This global feature is applied to pair and user.
      \item Length of active span, which is the number of days between first and last active day. This global feature is applied to pair, user and brand.
      \item Percentage of frequent users, where frequent means that a user has purchased the same brand more than once. This global feature is only applied to brand.
    \end{itemize}

\section{Blending and Ensemble}\label{sec:blending}

In this section, we first simply introduce our individual models, and then blending and ensemble approach for aggregating those models.

    \subsection{Individual Models}\label{subsec:model}

    We mainly use three types of individual model: regression, classification, and global scoring.
    For regression, we use Gradient Boosting Regression Tree (GBRT) as a single model.
    For classification, we use Logistic Regression (LR) and Random Forest (RF) as individual models.
    Fixed time span is applied to all these three models, and sliding time span is applied to GBRT and RF.
    The organizers of this competition provide distributed implementation of these classical learning models.
    For global scoring, we use a time decay function to directly score each instance:
    \begin{equation}
        score(user ID, brand ID) = \sum_N \sum_{T} \alpha_t \cdot \mathbb{I}(t) \cdot day^2(n),
    \end{equation}
    where $N$ is the total number of the instance actions, $T$ is the number of action types, $\alpha_t$ is the weight of action type $t$, $\mathbb{I}(t)$ is an indicator showing whether the action is of type $t$, and $day(n)$ is the day index of the $n$th action. Here $day^2(n)$ is a time decay factor.

    \subsection{Overall Framework}\label{subsec:framework}

\begin{figure}
    \centering
    \includegraphics[width=.47\textwidth]{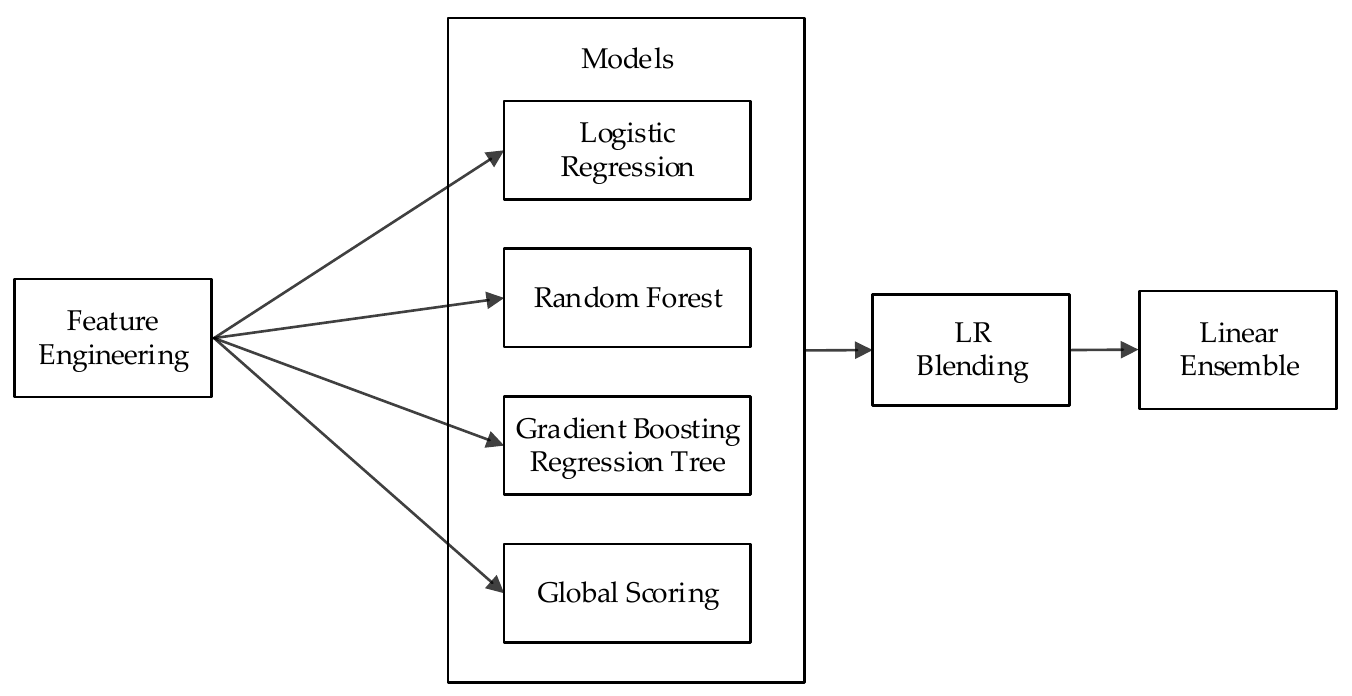}
    \caption{The overall framework of our proposed models.} \label{fig:framework}
    \end{figure}

    With the individual models, we use a two-stage approach for model blending and ensemble.
    We firstly use individual models to predict the training set, and apply Logistic Regression (LR) to blend those models group by group.
    Then, to avoid overfitting, we just use a simple linear ensemble to the blended models, and get our final prediction model.
    The overall framework is shown in Figure~\ref{fig:framework}.

    \subsection{Post Analysis}\label{sec:blending}

\begin{figure}
    \centering
    \includegraphics[width=.4\textwidth]{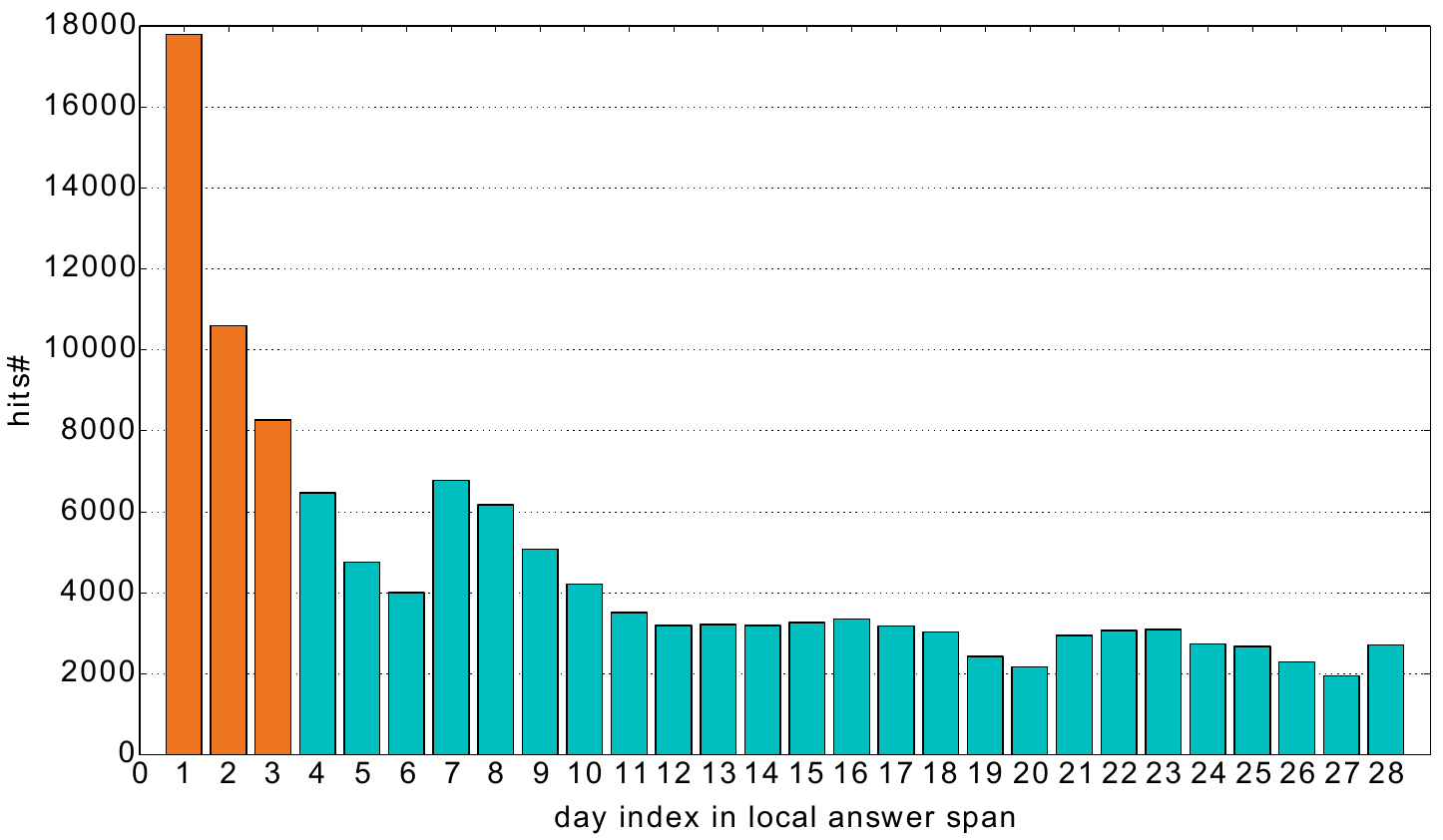}
    \caption{Day distribution of local prediction hits.} \label{fig:dist}
    \end{figure}

    We make some post analysis of our prediction result on local answer set to check the model performance in detail.
    We pick up all the hits in local prediction, and draw the day distribution of them, as is shown in Figure~\ref{fig:dist}.
    This analysis shows that our model are good at predicting user purchases in the first 3 days.
    It is clear to see that the predicting difficulty significantly increases after the first 3 days.
    This is understandable since useful information becomes weaker but noise becomes stronger in farther future.

\section{Conclusion}\label{sec:conclusion}

In this paper, we present our solution to Tmall Recommendation Prize 2014.
We construct instance for training and predicting in a time-dependent way, and model the purchase prediction task as a standard machine learning problem.
Two ways of time span construction are designed for generating feature and target of instance.
We mainly use three types of individual model: regression, classification, and global scoring.
We use a two-stage approach for model blending and ensemble, which effectively improve the prediction accuracy.


\end{document}